\documentclass[runningheads]{llncs}

\usepackage{graphicx}
\usepackage{amsmath,graphicx}
\usepackage{epstopdf}
\usepackage{array}
\usepackage{graphicx}
\usepackage{multirow}
\usepackage{color, soul}
\usepackage{cite}
\usepackage{url}
\usepackage{hyperref}
\usepackage{booktabs}
\usepackage{amsmath}
\usepackage{amssymb}
\usepackage{wrapfig} 
\usepackage{float} 
\usepackage[misc]{ifsym} 

\begin{document}

\title{A Deep Investigation of RNN and Self-attention for the Cyrillic-Traditional Mongolian Bidirectional Conversion\thanks{Rui Liu is the corresponding author.
This research is funded by the National Key Research and Development Program of China (No.2018YFE0122900), China National Natural Science Foundation (No.62066033), the High-level Talents Introduction Project of Inner Mongolia University (No. 10000-22311201/002) and the Young Scientists Fund of the National Natural Science Foundation of China (No. 62206136), Applied Technology Research and Development Program of Inner Mongolia Autonomous Region (No.2019GG372, 2020GG0046, 2021GG0158, 2020PT002), Young science and technology talents cultivation project of Inner Mongolia University (No.21221505), The Research Program of The National Social Science Fund of China (No.18XYY030).}}

%
%

\author{Muhan Na
\orcidID{0000-0002-8104-1749} \and
Rui Liu
$^{(\textrm{\Letter})}$
 \and
Feilong Bao
\orcidID{0000-0001-7312-1629} \and
Guanglai Gao
}

%
\institute{College of Computer Science, Inner Mongolia University, Hohhot, China\\
National \& Local Joint Engineering Research Center of Intelligent Information Processing Technology for Mongolian, Hohhot, China\\
Inner Mongolia Key Laboratory of Mongolian Information Processing Technology, Hohhot, China
\\
\email{Namuhan\_NMH@163.com, liurui\_imu@163.com, \{csfeilong, csggl\}@imu.edu.cn}}

\toctitle{RNN and Self-attention for the CTMBC}
\tocauthor{Muhan Na et al.}

\maketitle                

\begin{abstract}
Cyrillic and Traditional Mongolian are the two main members of the Mongolian writing system.
The Cyrillic-Traditional Mongolian Bidirectional Conversion (CTMBC) task includes two conversion processes, including Cyrillic Mongolian to Traditional Mongolian (C2T) and Traditional Mongolian to Cyrillic Mongolian conversions (T2C). 
Previous researchers adopted the traditional joint sequence model, since the CTMBC task is a natural Sequence-to-Sequence (Seq2Seq) modeling problem. 
Recent studies have shown that Recurrent Neural Network (RNN) and Self-attention (or Transformer) based encoder-decoder models have shown significant improvement in machine translation tasks between some major languages, such as Mandarin, English, French, etc. However, an open problem remains as to whether the CTMBC quality can be improved by utilizing the RNN and Transformer models.
To answer this question, this paper investigates the utility of these two powerful techniques for CTMBC task combined with agglutinative characteristics of Mongolian language. 
We build the encoder-decoder based CTMBC model based on RNN and Transformer respectively and compare the different network configurations deeply.
The experimental results show that both RNN and Transformer models outperform the traditional joint sequence model, where the Transformer achieves the best performance. Compared with the joint sequence baseline, the word error rate (WER) of the Transformer for C2T and T2C decreased by 5.72\% and 5.06\% respectively.

\end{abstract}

\section{Introduction}
Mongolian language belongs to the language group of the Altaic language family and is both the most widely spoken and most-known member of the Mongolic language family~\cite{book02}. The number of speakers across all its dialects may be 5.2 million, including the vast majority of the residents of Mongolia and many of the ethnic Mongol residents of the Inner Mongolia Autonomous Region of the People's Republic of China.
In Mongolia, the Mongolian character is currently written in Cyrillic Mongolian script~\cite{book03}.
In Inner Mongolia of China, the language is written in the Traditional Mongolian script~\cite{book02}.
The Cyrillic-Traditional Mongolian Bidirectional Conversion (CTMBC)~\cite{bao2014research} consists of Cyrillic Mongolian to Traditional Mongolian conversion (C2T) and Traditional Mongolian to Cyrillic Mongolian conversion (T2C). 
Therefore, the CTMBC facilitates the language communication between the compatriots of both countries and has great importance to the scientific, economic, and cultural fields of both countries.



The traditional method mainly focuses on the rule-based approach and the statistical model, such as the joint sequence model.
Specifically, Bao Sarina et al.~\cite{Sarina,li2011study} focused on the nouns and case suffixes translation in the CTMBC task and proposed a hybrid translation method that includes bilingual dictionary, rules
and N-gram language model.
Gao et al.~\cite{GaoMa2012} also proposed a hybrid method based on dictionaries and rules.
Feilong et al.~\cite{bao2014research} first adopted joint sequence model for the CTMBC task. In 2017, Feilong et al.~\cite{BAO2017} further proposed a hybrid method based on a combination of rule and joint sequence model. It adopts the rule-based approach to convert the words in the vocabulary, and uses joint sequence model to convert the out-of-vocabulary (OOV) words.

However, the conversion methods based on rule or statistical model perform some shortcomings. 1) The rule-based approach is unavailable when facing OOV words and loanwords~\cite{BAO2017}:
Mongolian words are made up of stems and suffixes. Different types of suffixes are added to the same stem to form different words, which leads to a huge vocabulary; Furthermore, the Mongolian language contains many loanwords that do not follow the Mongolian word-formation rules~\cite{book02};
2) The statistical model performs poor generalization ability and limited modeling capability: the joint sequence model holds shallow architecture and does not have strong nonlinear modeling ability~\cite{sutskever2014sequence,cho2014learning}.

We note that the CTMBC task can be interpreted as a standard machine translation task~\cite{bao2014research} between two languages. In other words, it is also a sequence-to-sequence (Seq2Seq)~\cite{sutskever2014sequence} modeling problem. The ``encoder-decoder'' structure~\cite{kalchbrenner2013recurrent,sutskever2014sequence,cho2014learning} have been successfully applied to various Seq2Seq tasks, including neural machine translation~\cite{sutskever2014sequence,bahdanau2014neural,wu2016google},
speech synthesis~\cite{wang2017tacotron,liu2020teacher,liu2018improving,liu2020exploiting,liu2020modeling},
and the grapheme-to-phoneme (G2P) conversion~\cite{bisani2008joint,yao2015sequence,wang2021joint}, etc. 
Recurrent neural networks (RNN) and self-attention based ``encoder-decoder'' models and pre-training language model~\cite{devlinetal2019bert,yang2019xlnet}
have lately received attention from the community~\cite{
cho2014learning,wu2016google,cho2014properties}.

For the CTMBC task, we don`t have enough data to train a large-scale pre-training model. Meanwhile, the RNN and Self-attention based encoder-decoder models have great performance in Seq2Seq modeling tasks. However, an open problem remains as to whether the CTMBC performance can be improved by utilizing the RNN and Transformer models. To answer this question, this paper investigates the utility of these two powerful techniques without large-scale pre-training model for CTMBC task combined with agglutinative characteristics of Mongolian language.
In this study, we validate the RNN and self-attention in the CTMBC task respectively. 1) We deep study the agglutinative characteristics of the Mongolian language, including the Traditional and Cyrillic Mongolian;
2) We build RNN and self-attention based CTMBC models according to the analyzed agglutinative characteristics; 
3) To identify the optimal network configurations, we investigate and compare the RNN and self-attention models with various configurations. The experimental results show that both RNN and self-attention models outperform the traditional joint sequence model, where the self-attention model achieves the best performance.

The main contributions of this paper include,
1) We conduct a deep investigation of RNN and self-attention for the CTMBC task;
2) We combine the agglutinative characteristics of Mongolian language with RNN and self-attention models to achieve outstanding performance.
To our best knowledge, this is the first deep investigation of the recent powerful deep learning models, including RNN and self-attention models, for the CTMBC task.

\vspace{-3mm}
\section{Task Challenges}
\vspace{-2mm}


\vspace{-1mm}
\subsection{Data Sparseness}
\textcolor{black}{
For machine translation tasks, a large amount of aligned sentence-level data is required.
Although this paper treats CTMBC as a word-level machine translation task, there is currently no high-quality and large-scale training data for model training due to the low-resource nature of Mongolian language. 
}

\vspace{-4mm}
\subsection{Agglutinative Characteristics}
\label{addchar}
For the Mongolian written form, there are two styles including Cyrillic Mongolian and Traditional Mongolian scripts. The similarity and difference between them in terms of agglutinative Characteristics can be summarized as follows. 

\subsubsection{Similarity} The similarity between Traditional and Cyrillic Mongolian scripts are summarized as word formation, pronunciation, and grammar rules.

    
    

\textbf{$\bullet$ Word formation:} 
Cyrillic Mongolian and Traditional Mongolian is an agglutinative language in which a new word is created by joining multiple suffixes to a word stem.
    
\textbf{$\bullet$ Pronunciation:} Cyrillic Mongolian can always find at least one Traditional Mongolian with the same pronunciation.
    
\textbf{$\bullet$ Grammar rules:} Cyrillic Mongolian retains most of the grammatical features of Traditional Mongolian. They are consistent in grammar.

\subsubsection{Difference} The difference between Traditional and Cyrillic Mongolian scripts are summarized as symbol systems, morphological rules and the correlation between pronunciation and spelling.

\textbf{$\bullet$ Different symbol systems:} Cyrillic script has 13 vowels, 20 consonants, 1 hardened character, and 1 softened character~\cite{BAO2017}. Traditional script has only 8 vowels and 27 consonants~\cite{BAO2017}. 
Cyrillic script performs case-sensitive and follows the rule of capitalization of the initial letter, Traditional Mongolian is not so. Traditional Mongolian have inconsistencies between code and presentation form. 
To avoid this phenomenon, in this work, we transform the Traditional Mongolian characters into their corresponding Latin transcriptions~\cite{lu2017language}.
    
\textbf{$\bullet$ Different morphological rules:} The morphological rules of Cyrillic Mongolian has 66 categories~\cite{Uganbater,dulamragchaa2017mongolian}, while Traditional Mongolian just has 4 categories~\cite{book02}. Such difference lead to the fact that the characters of the two kind of Mongolian words can not correspond one by one. For example, Traditional Mongolian word ``\includegraphics[width=0.23in]{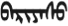}'' (Latin: bariyasv) has 8 characters, and its corresponding Cyrillic Mongolian word ``\includegraphics[width=0.35in]{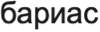}'' has 6 characters.

\textbf{$\bullet$ Different correlation between pronunciation and spelling:} The pronunciation and spelling of Cyrillic Mongolian are one-to-one correspondence, Traditional Mongolian is otherwise, occur a one-to-many relationship. 
In Traditional Mongolian, the vowels or consonants may be dropped, added, or changed when reading. As a result, one Cyrillic Mongolian word may correspond to more than one Traditional Mongolian word 
. For example, Cyrillic Mongolian for both ``\includegraphics[width=0.18in]{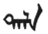}'' (Latin: dala, means: expand) and ``\includegraphics[width=0.18in]{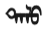}'' (Latin: dalv, means: shoulder) is ``\includegraphics[width=0.18in]{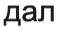}''.



\textcolor{black}{Mongolian words are constructed by successively concatenating suffixes to stems, which results in the word forms in Mongolian being very large and having a lot of out-of-vocabulary words. Meanwhile, different morphological rules make it difficult for the model to learn the mapping relationship between two Mongolian characters.}

{
The above data sparseness and agglutinative characteristics problems bring huge challenge for our CTMBC task. The agglutinative characteristics of Traditional and Cyrillic Mongolian scripts are the knowledge we must master for the CTMBC task. In the next section, we will fully explore the above language knowledge to complete the C2T and T2C conversion.}

\vspace{-2mm}
\section{RNN and Self-attention based Encoder-Decoder Frameworks for C2T and T2C}
\vspace{-1mm}
In this section, we will introduce the overall Encoder-Decoder frameworks for C2T and T2C, at first. Then we will explain their workflows in detail respectively.
Last and not least, the backbone of the encoder-decoder framework, including RNN and self-attention, will be introduced.

\begin{figure}
\includegraphics[width=\textwidth]{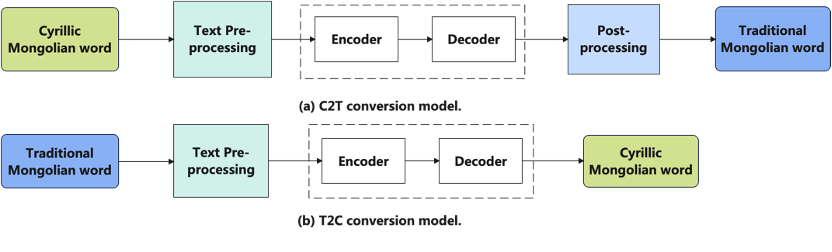}
\vspace{-2em} 
\caption{The overall frameworks of (a) C2T conversion model, (b) T2C conversion model. \textcolor{black}{Our task is similar to G2P task~\cite{wang2021joint}. In the process of T2C conversions, our input is Traditional Mongolian word and output is Cyrillic Mongolian word, while vice versa in C2T conversion.}} \label{fig2}
\end{figure}

\vspace{-2mm}
\subsection{Overall Framework}
\vspace{-1mm}
As shown in Fig.~\ref{fig2}, the overall frameworks of C2T and T2C are illustrated in the upper and bottom panels respectively.
\textcolor{black}{We follow G2P~\cite{wang2021joint} workflow to build our C2T and T2C frameworks, in which the input and output both are word-level.} 
As mentioned in Section.~\ref{addchar}, there are some similarities and differences between Cyrillic and traditional Mongolian scripts in terms of agglutinative Characteristics. Therefore, we make necessary processing for Mongolian scripts and design the workflows for C2T and T2C specifically.

\vspace{-2mm}
\subsection{Workflow of C2T}
\vspace{-1mm}
\textcolor{black}{
As shown in Fig.~\ref{fig2}(a), C2T conversion consists of 3 parts, including ``Text Pre-processing'', ``Encoder-Decoder'' and ``Post-processing''. 
Text Pre-processing takes the Cyrillic Mongolian word as input to output the character sequence. Then the encoder reads the character sequence to generate the high-level hidden representation, which is then fed to the decoder to predict the Latin transcriptions of Traditional Mongolian characters. At last, Post-processing module converts the Latin transcription to the Traditional Mongolian scripts as the final conversion results.
}

\vspace{-2mm}
\subsubsection{Text Pre-processing}
\label{textpp}

Given that $C$ is the input Cyrillic Mongolian word, Text Per-processing can output the character sequence $\textbf{c}=c_{1},c_{2},c_{3},c_{4},...,c_{n}$ ($n$ means the sequence length or character number).
Words and sub-words also are two possible training units. However, the complex word or sub-word generation pipeline 
will bring some noise and bring about side effects. In other words, incorrect tokenization or separation results will affect model performance.
\vspace{-3mm}
\subsubsection{Encoder-Decoder}
\label{encdec}
The encoder summarizes the input sequence, Cyrillic Mongolian characters, into a set of vectors while the decoder conditions the encoded input sequence, and generates the output sequence, Traditional Mongolian Latin transcription characters, one token at a time.
Let $\textbf{o}=o_{1},o_{2},...,o_{m}$ be the sequence of $m$ symbols in The encoder is simply a function of the following form: $x=Encoder(c_{1},c_{2},...,c_{n})$, in which $\textbf{x}=x_{1},x_{2},...,x_{m}$ is a list of fixed size vectors.


The decoder is often trained to predict the next word $o_{t}$ ($t \in [1,m]$) given all the previously predicted words $o_{1},o_{2},...,o_{t-1}$. In other words, the decoder defines a probability over the conversion $\textbf{o}$ by decomposing the joint probability into the ordered conditionals:
$P\left(o\middle|x\right)=\prod_{t=1}^{M}P\left(o_t\middle| o_{1},o_{2},...,o_{t-1};x_{1},x_{2},...,x_{n}\right)$.



\subsubsection{Post-processing}
The output of decoder is character-level Latin transcription. Therefore, Post-processing aims to restore the Latin transcription to Traditional Mongolian word $T$.
\subsection{Workflow of T2C}
\vspace{-2mm}
As shown in Fig.~\ref{fig2}(b), different from the C2T conversion process, T2C conversion consists of the following 2 parts, including ``Text Pre-processing'' and ``Encoder-Decoder''.
Text Pre-processing module aims to convert Traditional Mongolian word to their character-level Latin transcription.
The Encoder processes the character-level Latin transcription and outputs the high-level representation. Decoder module predicts the Cyrillic Mongolian as the result.
\vspace{-4mm}
\subsubsection{Text Pre-processing}
There are two steps in the Text Pre-processing of T2C. First, convert the Traditional Mongolian into Latin transcription.
Second, divide word-level Latin transcription into character-level Latin sequence.
\textcolor{black}{As mentioned in Section~\ref{addchar}, in this work, we transform the Traditional Mongolian characters to their corresponding Latin transcription for model training.}
Similar with Section~\ref{textpp}, we also divide Latin transcription into Latin characters.
\vspace{-4mm}
\subsubsection{Encoder-Decoder}
In the T2C task, the encoder-decoder module is the same as that, as described in Section~\ref{encdec} in the C2T task. The only difference is that the input of T2C task is Traditional Mongolian Latin transcription characters and the output is Cyrillic Mongolian characters.

\vspace{-4mm}
\subsection{Backbone of Encoder-Decoder Framework}\
\vspace{-1mm}
In this section, we use C2T as an example to introduce the backbone of Encoder-Decoder Framework. To validate the RNN and Self-attention for the CTMBC task, we employ attention based RNN and self-attention model as the backbone of Encoder-Decoder framework. We will introduce the details next.

\vspace{-4mm}
\subsubsection{RNN}
\label{rnn}

Specifically, RNN-based encoder is implemented with a multi-layer bidirectional LSTM (BiLSTM), which transforms the input Cyrillic Mongolian sequence $c$ into a high-level representation $h=\left\{h_n\right\}_{n=1}^N $. 
The RNN-based decoder takes $h $ and all previously predicted outputs $o_{1:m-1} $ as input, producing the probability distribution of the token $o_m$.
Specifically, at each decoder time step $t$, the posterior distribution of the predicted output is generated from the cascade of decoder states st and context vector $a_t $. The $a_t $ is the context information produced by the attention module based on the hidden state of the encoder and decoder. 






\vspace{-4mm}
\subsubsection{Self-attention}


Unlike the RNN-based encoder-decoder framework, the self-attention based encoder-decoder framework, that is  Transformer, replaces the RNN modules with the pure self-attention mechanism.
Specifically, Transformer encoder consists of $N$ identical Transformer blocks~\cite{vaswani2017attention}. Each block consists of two sub-layers, including the multi-head self-attention mechanism and the fully connected feed-forward network. Residual connection and normalization are added to each sub-layers. 



Transformer decoder also consists of $N$ identical Transformer blocks~\cite{vaswani2017attention}, which include masked multi-head self-attention, multi-head self-attention and the fully connected feed-forward network. Residual connection and normalization are also added to each sub-layers.
In addition, different from the RNN based model, the Transformer encoder and decoder take a position encoding~\cite{vaswani2017attention} as an additional input.


\vspace{-3mm}
\section{Experiments and Results}
\vspace{-2mm}
\subsection{Datasets}
\vspace{-2mm}
We report the experiments on a Cyrillic-Traditional Mongolian mapping dictionary, denoted as ``\textit{Mon\_data\_63668}''.
The \textit{Mon\_data\_63668} dataset includes 63668 word pairs which collected from the New Mongolian-Chinese Dictionary~\cite{book01}.
Note that 58436 pairs were randomly selected as the training set, and the remaining 5232 word pairs were used as the model testing set.

\vspace{-4mm}
\subsection{Evaluation Setup}
\vspace{-3mm}
\subsubsection{Comparative study}
\textcolor{black}{
We implement three frameworks for C2T and T2C in a comparative study.
RNN and self-attention based encoder-decoder models are studied for the first time for CTMBC, while Joint sequence model is the baseline and re-implementation of~\cite{bao2014research}.}

\textbf{$\bullet$ Joint Sequence Model (``Joint'' for short) (Baseline):}
    The idea of the joint sequence model~\cite{bisani2008joint} is to represent the relationship between an input sequence and an output sequence in terms of a common sequence of joint units composed of input and output symbols. \textcolor{black}{N-gram language model is used to predict character during decoding.}
    Note that we will seek the optimal model configuration by adjusting the value of $N$ in N-gram.\textcolor{black}{ We infer the model parameters through the expectation maximization and trim the evidence to avoid over-fitting. We also discount evidence for smoothing.}


\textbf{$\bullet$ RNN based Encoder-Decoder (New)}
We implement two architectures, including RNN based Encoder-Decoder without attention (``RNN'' for short) and the attention based RNN Encoder-Decoder as mentioned in Section~\ref{rnn} (``RNN+ATT'' for short), for comparison. Note that we set the RNN hidden size to 512 and 1024, and set the hidden layer to 1, 2, and 4 to explore the optimal model configuration.

\textcolor{black}{For RNN training, the input of encoder is 128-dimensional character sequence and the output of decoder is mongolinan characters. The parameters of RNN are set as following: batch size = 32, epochs = 100, learning rate = 0.0005. We decrease the learning rate every 20 epochs by a factor of 0.9. The loss function is using the cross-entropy(CE) criterion. We conduct experiments on an NVIDIA GPU (Tesla P40).}

\textbf{$\bullet$ Self-attention based Encoder-Decoder (``Transformer'' for short) (New):}
``Transformer-Tiny'' model~\cite{vaswani2018tensor2tensor} to build our self-attention based Encoder-Decoder model since it holds a smaller vocabulary than ``Transformer-base''~\cite{vaswani2018tensor2tensor} model and therefore is more suitable for our task. Note that we set the attention head to 2 and 4, and set the hidden layer to 1, 2, 4, and 6 to explore the optimal parameter combination.

\textcolor{black}{All the comparative experiments are conducted in the Tensor2Tensor~\cite{vaswani2018tensor2tensor}. We set embedding dimensional = 128, train steps = 100000, batch size = 4096,  learning rate = 0.2 and the learning rate warm-up~\cite{popel2018training} steps to 8000.  All word embeddings are initialized randomly and then updated with the whole model. We conduct experiments on an NVIDIA GPU (Tesla P40).}

\vspace{-4mm}
\subsubsection{Metrics}
The result is performed in terms of word error rate (WER) and character error rate (CER). The formulas of WER and CER are as follows: $WER=1-\frac{N_{corrent}}{N_{total}}$, $CER=\frac{N_{ins}+N_{del}+N_{sub}}{N_{charatertotal}}$.
$ N_{corrent}$ is the number of correctly predicted Mongolian words; $ N_{total}$ is size of test set; $ N_{ins}$ is the number of character insertion errors; $ N_{del}$ is the number of character deletion errors; $ N_{sub}$ is the number of character substitution errors; $ N_{charatertotal}$ is the number of word characters.

Following~\cite{bisani2008joint,yao2015sequence} in the case of multiple references Mongolian, the variant with the smallest edit distance is used. Similarly, if there are multiple references of Mongolian for a word, a word error occurs only if the predicted Mongolian doesn’t match any of the references.



\begin{figure}[th]
\centering
\vspace{-1.3em}
\includegraphics[width=\textwidth]{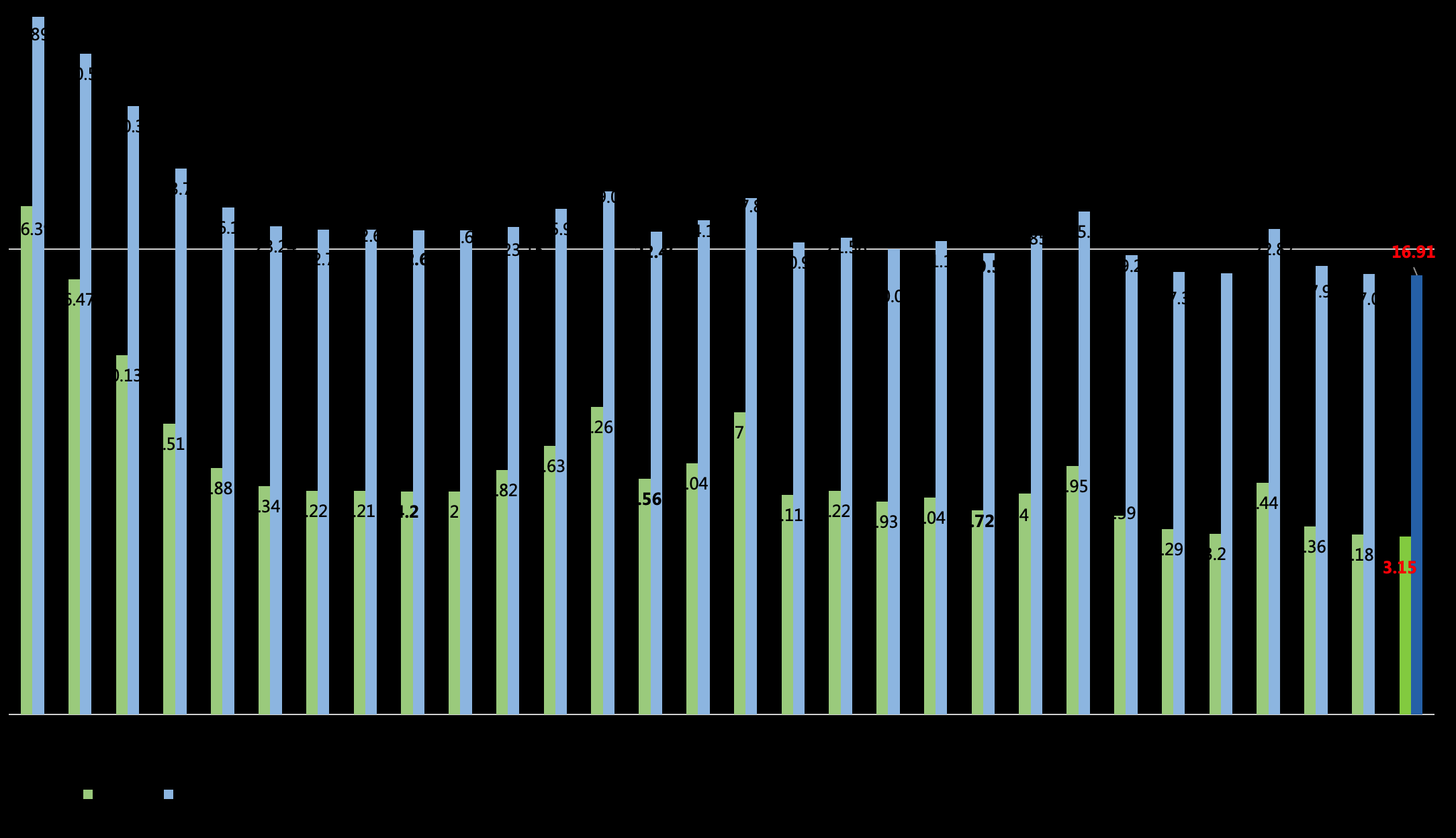}
\vspace{-2em} 
\caption{The WER and CER results of C2T. The first panel is results of Joint sequence model and ``N'' is delegates N of N-gram. The second panel and third panel is results of RNN based model and RNN+ATT based model, ``U'' represents BiLSTM hidden units and ``L'' represents BiLSTM hidden layer. The fourth panel is results of Transformer model under different parameters, ``S'' represents the layer size of Transformer and ``H'' represents the attention head number.} \label{fig6}
\vspace{-2.em}
\end{figure}

\begin{figure}[th]
\centering
\includegraphics[width=\textwidth]{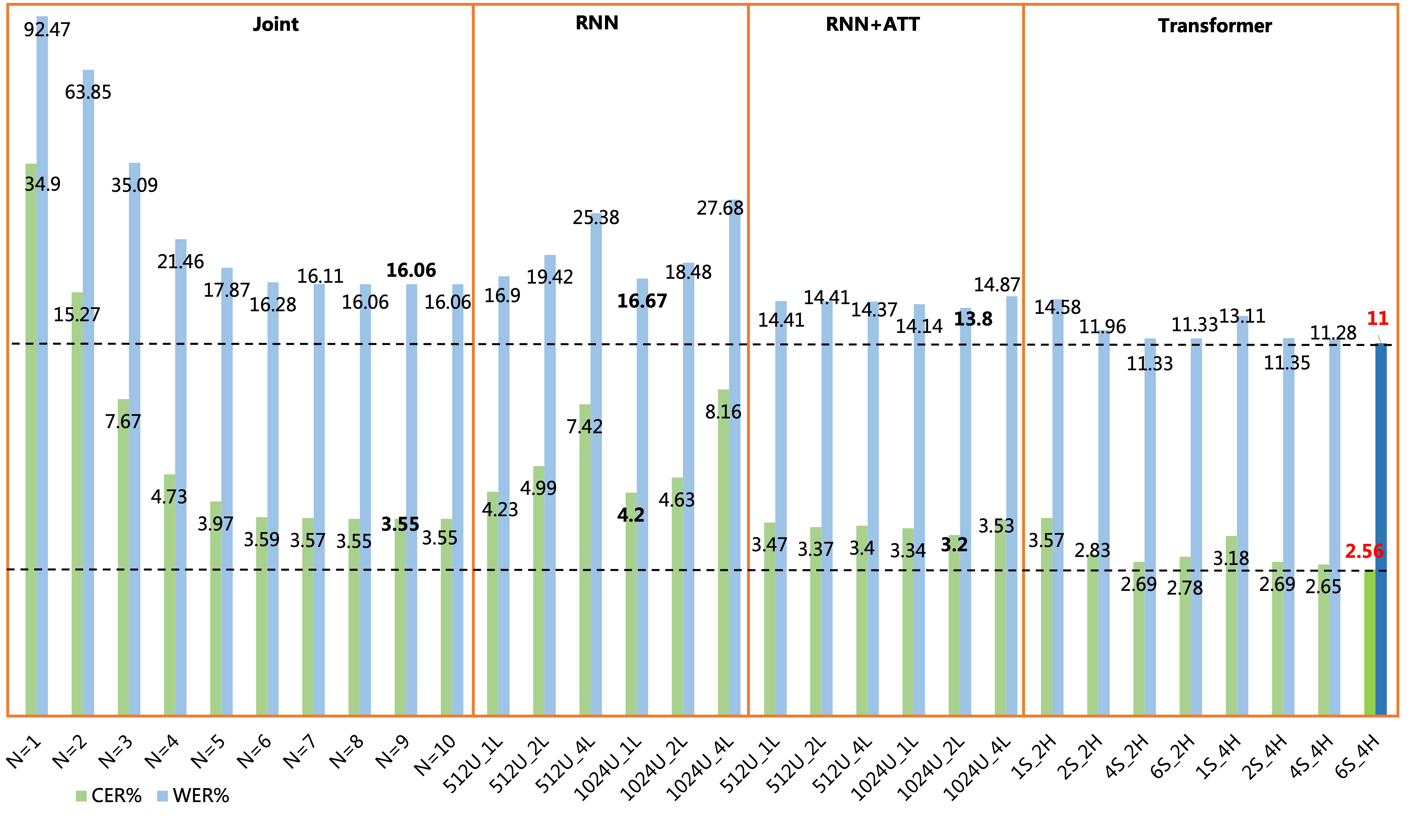}
\vspace{-3em} 
\caption{The WER and CER results of T2C. 
} \label{fig7}
\vspace{-1.5em} 
\end{figure}

\vspace{-4mm}
\subsection{Performance Comparison for C2T and T2C}
\vspace{-2mm}

Fig.~\ref{fig6} shows the WER and CER results of all models under different parameters for C2T. We will analysis the results next.

\vspace{-2mm}
\subsubsection{Joint vs. (RNN \& RNN+ATT)}

As shown in the first panel of Fig.~\ref{fig6}, by increasing $N$ from 1 to 10, we can find that the WER and CER are basically decreasing gradually.
``N=9'' for optimal performance, with with WER=22.63\% and CER=4.2\%.


From the second panel of Fig.~\ref{fig6}, the 1-layer BiLSTM with 1024 hidden units obtains the best performance, which can reach 4.56\% on CER and 22.46\% on WER. 
The optimal result, RNN(1024U\_1L), is almost the same as ``N=9''. We also fix hidden layer number of BiLSTM and adjust the hidden unit size to observe the performance. It can be found that the performance of the model can be improved by increasing the hidden unit size. For example, RNN(512U\_1L) outperforms the RNN(1024U\_1L) model. 
Stacking more hidden layers sometimes does not lead to significant improvements. For example, RNN(1024U\_1L) beats the RNN(1024U\_4L) model.

In the third panel of Fig.~\ref{fig6}, we can find that the 2-layer BiLSTM with 1024 hidden units, RNN+ATT(1024U\_2L), obtains the best performance, which can reach 3.72\% in CER and 19.51\% in WER.
Compared with the optimal results of RNN, RNN(1024U\_1L) model, the WER of RNN+ATT(1024U\_2L) is reduced by 2.95\%. 
We believe that adding the attention mechanism to the RNN based encoder-decoder model can effectively capture the  internal information of input Mongolian character sequence better and improve the model performance.

T2C has similar results as C2T. As shown in Fig.~\ref{fig7}, RNN(1024U\_L1) model outperforms the ``N=8'' model, RNN+ATT(1024U\_L2) model beat the ``N=8'' model, which consists of C2T.

\vspace{-4mm}
\subsubsection{Joint vs. Transformer}
The fourth panel of Fig.~\ref{fig6} show the WER and CER of Transformer model under different parameters.

The results show that the Transformer framework achieves best performance when layer size is 6 and attention head is 4.
``6S\_4H'' achieves 3.15\% on CER and 16.92\% on WER and gains 5.72\% relative reduction compared to the ``N=9'' model in terms of WER.
We also find that increasing the attention head number can effectively improve the model performance. For example, the ``6S\_4H''  and the ``4S\_4H'' outperforms the ``6S\_2H'' and the ``4S\_2H'' respectively.

The result of T2C is similar to C2T. In Fig.~\ref{fig7}, ``6S\_4H'' model outperforms the ``N=8'' model, which has identical conclusion to C2T.

\vspace{-4mm}
\subsubsection{(RNN \& RNN+ATT) vs. Transformer}
Comparing the optimal results of RNN and Transformer, Transformer achieves better performance.
\begin{sloppypar}
\textcolor{black}{For example, the ``6S\_4H'' achieves 16.92\% WER and outperforms RNN(1024U\_1L) with 5.54\% WER.
The self-attention mechanism has stronger ability to capture long-term dependencies within input sequence, therefore brings better performance.
Comparing RNN+ATT(1024U\_2L) and ``6S\_4H'', ``6S\_4H'' beat RNN+ATT(1024U\_2L) with 2.95\% WER. 
The reason is that the self-attention mechanism can capture global context information, it is helpful for model prediction.
}
\end{sloppypar}

The results of T2C resemble C2T. As shown in Fig.~\ref{fig7}, ``6S\_4H'' model beat the RNN(1024U\_1L) and RNN+ATT(1024U\_2L), which is consistent with C2T.


\subsection{Case study}

\begin{figure}
\centering
\vspace{-1.5em} 
\includegraphics[width=\textwidth]{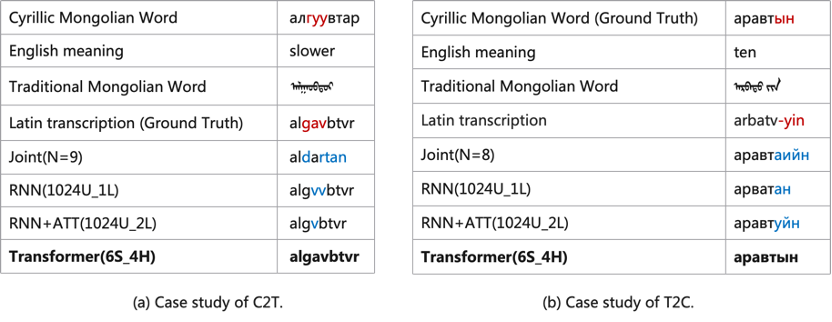}
\vspace{-2em} 
\caption{The example of translated words produced by various models. The Latin characters not correctly translated are highlighted in blue. The long vowels and suffix are highlighted in red. 
} \label{fig8}
\vspace{-2em} 
\end{figure}

To analyze the translated words produced by the various model, we use two examples extracted from test set to compare them in Fig.~\ref{fig8}. The  wrong translated characters are highlighted in blue.

\textcolor{black}{
As shown in Fig.~\ref{fig8}(a), we take a Cyrillic Mongolian word ``\includegraphics[width=0.5in]{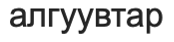}'' as an example to show the C2T conversion results.The Latin transcriptions is ``alagavbtvr''. Note that there is vowel substitution phenomenon in the pronunciation of this word. The traditionally Mongolian word is pronounced with the fifth vowel ``a'' becoming ``v'' and the ninth vowel ``v'' becoming ``a''.  The fifth row shows the conversion result of Joint model. We can find that there are 4 wrong characters in the results, which performs poor performance, while RNN and RNN+ATT models just have 2 and 1 wrong characters respectively. We further observe that the translated word output from Transformer model exactly the same as the target Latin transcriptions ``alagavbtvr'' and contains no errors.}

Similar with Fig.~\ref{fig8}(a), as shown in the Fig.~\ref{fig8}(b), we also take a traditional Mongolian word ``\includegraphics[width=0.35in]{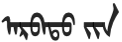}'' as an example to show the T2C conversion results. The Latin transcriptions is ``arbatv-yin''. We observe the Joint, RNN, and RNN+ATT models all have some wrong conversion characters, while Transformer model contains no errors. Note that the Transformer improves the conversion accuracy of long vowels and suffix that are highlighted in red.

The case study results show that the RNN and Transformer are stronger than Joint model and Transformer achieves the best performance, which is consistent with our previous results.

\vspace{-1em}
\section{Conclusion}
\vspace{-1em}
In this paper, we introduce RNN and self-attention based models into the Cyrillic-Traditional Mongolian Bidirectional Conversion (CTMBC) task. This is the first deep investigation of the recent powerful deep learning models for the CTMBC task.
Compared with the joint sequence model baseline, RNN based models and self-attention model gains significant improvements. Note that self-attention model achieves the best performance at both T2C and C2T tasks. In addition, we also compared the parameter setting of each model, such as hidden layer number, hidden unit size, attention head number, etc., in detail to determine the optimal model configuration.
Note that this paper mainly studies the CTMBC task at word-level, which similar with G2P. However, word level conversion is plagued by frequent polyphonic words and lack of contextual information, which may limit model performance. In the future work, we will focus on CTMBC task at sentence-level to make full use of context information for more accurate conversion performance.

%
%
%
%


\end{document}